# Stylomech: Unveiling Authorship via Computational Stylometry in English and Romanized Sinhala

Nabeelah Faumi[1], Adeepa Gunathilake[1], Benura Wickramanayake[1], Deelaka Dias[1], TGDK Sumanathilaka[2]
[1]Informatics Institute of Technology, Colombo 006, Sri Lanka
[2]School of Mathematics and Computer Science, Swansea University, UK
Corresponding author: faumi.20222404@iit.ac.lk

*Abstract*— With the advent of Web 2.0, the development in social technology coupled with global communication systematically brought positive and negative impacts to society. Copyright claims and Author identification are deemed crucial as there has been a considerable amount of increase in content violation owing to the lack of proper ethics in society. The Author's attribution in both English and Romanized Sinhala became a major requirement in the last few decades. As an area largely unexplored, particularly within the context of Romanized Sinhala, the research contributes significantly to the field of computational linguistics. The proposed author attribution system offers a unique approach, allowing for the comparison of only two sets of text: suspect author and anonymous text, a departure from traditional methodologies which often rely on larger corpora. This work focuses on using the numerical representation of various pairs of the same and different authors allowing for, the model to train on these representations as opposed to text, this allows for it to apply to a multitude of authors and contexts, given that the suspected author text, and the anonymous text are of reasonable quality. By expanding the scope of authorship attribution to encompass diverse linguistic contexts, the work contributes to fostering trust and accountability in digital communication, especially in Sri Lanka. This research presents a pioneering approach to author attribution in both English and Romanized Sinhala, addressing a critical need for content verification and intellectual property rights enforcement in the digital age.

## I. Introduction

Authorship Attribution is also known as authorship identification or authorship verification. It is the process of attributing or identifying the author of a specific text, using lexical, semantic and syntactic features[1]. Authorship verification (AV) addresses the problem of determining whether a given text was written by a certain author A or not[2]. Stylometry is a key element of authorship attribution. It involves the analysis of writing style to pinpoint writers. It considers different linguistic patterns and features in texts, including sentence structure, vocabulary choices and other stylistic components[3]. Many models exist for English Authorship attribution using various technologies. Lee *et al.* [4] used a Random Walk model to understand if the authors of 2 text messages were the same. Furthermore, DNNs in authorship attribution were also used to identify authors of fraudulent activity[5]. Iyer *et al.* use various techniques from regression and SVM models to train the works of 50 different authors which can predict the author of the new text [6]. This, however, only worked for a particular set of 50 authors and doesn't allow for a more general approach. The use of cosine similarity especially when it comes to multilingual text alongside feature extraction was done by Anwar et al [7] which was evaluated for both Urdu and English. However, the model was limited to a domain, since non-domain-specific texts were misclassified. Previous research in authorship attribution has predominantly centred on classifying text from a predefined set of known authors using machine learning techniques. These limitations in the current approaches show the importance of an advanced model which could be able to identify the suspect author, by only referring to a limited number of resources.

Further author verification models present today's work by training intensely on one specific author and must be retrained every time. The presented work combats this by only considering 2 pieces of text to verify the author, by training on if 2 pieces of text are authored by the same person. Romanized Sinhala is considered to be a low-resourced language, and a very limited amount of research is conducted on the author's identification using natural text. Previous research in converting transliterated text to Sinhala and Romanized Sinhala text analysis has been conducted [8].

TABLE I. ANALYSIS OF ROMANIZED SINHALA CONTEXT

| Sinhala script | මම ගෙදර ගියා |
|---|---|
| Latin script | Mama gedara giya |

Table 1 represents the form of usage of English alphabets to present Sinhala characters in the Sri Lankan context. This form of communication is prevalent within the Young Adult population of Sri Lanka because Latin scripts are the most common scripture system used in computer input systems. However, no previous research in the area of stylometric analysis has been explored for Romanized Sinhala. According to our knowledge, this is the first study which has been conducted on stylometry analysis in the Romanized Sinhala context.

Based on these gaps, the key research questions of this study are as follows:
(1) Can authorship attribution models be effectively developed for Romanized Sinhala text?
(2) Can such a model generalize across different domains and linguistic structures?

To address these questions, a novel Romanized Sinhala authorship attribution model was developed, leveraging textual data and stylometric features. The model demonstrates the ability to classify authorship in both English and Romanized Sinhala text, providing a lead on stylometric analysis for Romanized versions of other languages. This proposed solution tries to tackle the problems of data scarcity issue during the inference process, by introducing the stylometric information of suspected text and the author's known text only to predict the suspected text by the author or not.

The major contribution of the studies can be listed as follows.
- Introducing a novel English and Romanized Sinhala Authorship Attribution model based on stylometry feature analysis.

This section provides a comprehensive understanding of the problem, highlights the existing gap in the research, and outlines the contributions of this study. Moving forward, the paper will discuss the related works, present the methodology employed, and analyze the results. The discussion will also cover potential future works and directions for further research.

## II. RELATED WORKS

In the past few years, copyright claiming and author identification on anonymized content have been highly discussed and explored. However, the approaches suggested are computationally expensive or only limited to English and a fixed corpus of authors. There still exists a large gap in the low-resource's languages like Sinhala in the domain which needs extensive work to be performed.

Much research regarding applying stylometric analysis for attributing authors to anonymous or pseudo-anonymous documents has been done. Have been analyzing the efficiency of using n-gram analysis for stylometry. Their work revealed that bigram, trigram and quadgram are reliable metrics for stylometric analysis for novels [9]. Also, for the text extracted from novels, they found that a correlation exists with the usage of apostrophes, adjectives and dashes. In the research, hapax legomena and Yule's I vocabulary richness measure have been used as input features. The term "hapax legomena" means words that have only one occurrence in the given text. Also, character frequencies, n-gram frequencies and function word frequencies have been used as input features. The frequency of word lengths and frequency of word shapes have also been used [10]. From the previous research, features such as type-token ratio, mean word length, mean and the standard deviation of sentence length, mean paragraph length and syntactical features such as number of commas per thousand tokens and the number of "and"/"or" per thousand tokens were identified as mainstream methods for stylometric analysis[11].

In academic and professional settings stylometry models are being used to preserve academic integrity and detect plagiarism. A model proposed by Hoshiladevi Ramnial et al. was analysed with the Sequential Minimal Optimization (SMO) algorithm to detect plagiarism among 5 authors, and the model accurately identified cases where authors had plagiarized their work, as inconsistencies in stylometry were evident. This was confirmed by the Turnitin database, which identified high levels of plagiarism in the author's work [12]. From cyber security and criminal cases, a plethora of work has been carried out in the field, A paper used Random Walk Features from Chat Messages to detect phishing messages based on stylometric analysis of the message being sent and the previous chat logs, this resulted in an 86% accuracy rating for verifying the user [4]. In the paper, stylometry has been applied to identify cybercriminals in darknet forums. In the research, it was possible to identify authors of text written by cybercriminals using l33t-speak style conversations with an accuracy level of 77% to 84%. Hence, providing a successful demonstration of applying stylometric analysis to a cybercrime context [13]. Further research on Deep Learning for pattern recognition in stylometric analysis has been used, A deep learning model based on Stacked Denoising Autoencoders (SDAE), this experiment was done by using 50 authors(100 documents from each). The system(SDAE) achieves a remarkable classification accuracy of up to 95.12%, surpassing the accuracy of around 80% achieved by previous studies using the same dataset and feature sets, this was achieved using support vector classification [14]. Further, a study focusing on identifying authors of emails used Deep Neural Networks (DNN) and model-based clustering techniques averaging an accuracy of over 90%.These techniques and technologies have significantly advanced the field of stylometry and have promising applications in authorship attribution in areas ranging from digital forensics to plagiarism detection [5].

The use of various machine learning techniques for authorship attribution is a booming research topic that has been used multiple times with the use of SMOs, SVMs, k-NNs, Naïve Bayes and regression methodologies all having relatively high accuracy but they are limited when it comes to the fact that it can only be applied to authors in their datasets [6]. Further an author verification the Imposter method which involves collects impostor texts—samples not authored by the reference author—using search engines, creating a diverse pool of writing from multiple authors, this helps with identifying if the work doesn't belong to the suspect author based on its relation to author impostor texts[15]. Profile based author identification plays a big role in authorship verification and attribution, which determines specific attributes belonging to an author as its basis of verification, this may help narrow down and can act as features when it comes to author attribution [16]. Pertaining to multi-language AA, a study used Romanian

BERT model to get rich, context-based word representations, which capture the meaning of words in different sentences. These representations are combined with important language features, like word frequencies and grammar points for the Romanian language. This hybrid transformer method worked wonders to help identify features in languages, where resources aren't heavily available with an F1 score of 0.87[17].

When we look at works pertaining to Romanized Sinhala, the results are quite limited, In this work reverse transliteration is used to convert a Sinhala text written in Latin scripts to Sinhala scripts using the Swa-Bhasha Romanized Sinhala reverse transliteration model [8]. The model achieves 84% word-level accuracy and 93% character-level accuracy. Further, an authorship attribution model is present for Sinhala in intra- and cross-topic authorship attribution settings., where they use stylometry to evaluate the relevant semantic and syntactic features in Sinhala, that can be applied to multiple different languages as well. The findings here suggested that this was set to a specific author corpus whereas when the corpus increases the accuracy decreased, they concluded that kNN serves as the best model with an accuracy of 86% [18]. Looking at the above findings it is evident that Stylometry and Author attribution encompasses a variety of fields and disciplines. This makes it so that the techniques used for stylometric comparisons between texts are diverse, and ever-evolving. While most models seem to focus on using linear models and SVM for classification and comparison, it is important to note that there is a range of algorithms being employed. The algorithms' effectiveness also highly relies on the features used for stylometric analysis and directly plays into the accuracy and usability of many frameworks, and models for authorship attribution. This synergy is what shapes the land of authorship attribution in the context of machine learning and stylometric analysis.

III. METHODOLOGY

A. Dataset Creation

This model aims to be a general-purpose model, as a result the model isn't trained on textual data, but rather the similarity scores for each feature, between 2 texts. This ensures the model need not be re-trained and also ensures that the model is applicable to texts outside of the trained dataset as the values are only looked at, instead of the actual textual data. This allows for a more robust approach, that hasn't been done before.

1) English

The dataset was constructed by extracting features from publicly available "blog" and "tweets" datasets from which a combination of 100 different random authors. These data sources were cleaned to remove any identifying facts about the author, and further other pre-processing steps, like removing picture based emojis were removed. From each corpus, two pieces of text were selected, and their similarities were computed. These similarity scores were then compiled into a dataset, where a value of 1 was assigned if the two texts were authored by the same individual, and 0 otherwise, this served as the basis for the model training.

Following the dataset compilation, a cleaning process ensued to enhance its quality. To ensure data integrity and avoid redundancy, all duplicate entries were systematically eliminated from the dataset. Features exhibiting a variance of less than 0.05 were identified and subsequently removed to streamline the dataset and mitigate dimensionality concerns. Only features surpassing this variance threshold were retained for further analysis and modelling, there were 883 separate instances once cleaned.

2) Romanized Sinhala

WhatsApp chats exported from 50 different volunteers' conversations were used. Sensitive messages were removed and the order of each text message in each chat was scrambled to make the conversations meaningless in order to protect volunteer privacy. Then chats were split into small chunks.

Pictures, emojis and GIFs were removed from each chat systematically, so the analysis could be purely done using textual context clues. Then a pair of such chunks were taken at once and features, namely, the English to Sinhala ratio of each chunk, text edit distance between similar Romanized Sinhala words between the two chunks were extracted and saved to a csv file with a column identifying if the two chunks used are belonging to the same author or not. This way of dataset generation and pre-processing allowed the model to handle user input regardless of size or even textual context and is able to understand patterns based off of 2 texts, instead of a text corpus.

B. Feature Extraction

We used standard stylometric features for feature extraction, commonly used features like POS tagging, mean sentence length, and N-gram analysis.[11]

1) Feature Extraction for English Model

The text from the English dataset underwent tokenization, a process that breaks it down into individual words, punctuation marks, and symbols, enabling the extraction of meaningful features for further analysis. Following tokenization, Part of Speech (POS) tagging was applied using the Natural Language Toolkit (NLTK), assigning each token its respective part of speech (e.g., noun, verb, adjective). This tagging aids in capturing syntactic patterns and linguistic nuances. Additionally, active and passive voice detection was performed to identify sentence structures, providing insight into the author's preferred writing style. The frequency of punctuation marks, such as commas and periods, was calculated to assess sentence length, complexity, and rhythm. Similarly, the distribution of sentence lengths, measured in word count, was analyzed to uncover patterns in sentence structure and complexity.

Function words, which serve grammatical rather than lexical purposes, were tagged, and their frequency was calculated to further understand the structural aspects of the text. An N-gram transition graph was also generated, analyzing sequences of adjacent words to depict the probability of word sequences in the text, offering insights into language usage patterns. Finally, a gender prediction model based on Long Short-Term Memory (LSTM) was implemented using the "md_gender_bias" dataset, demonstrating high accuracy in predicting gender from linguistic patterns and contextual cues.

*2) Feature Extraction for Romanized Sinhala Model*

A script was developed to calculate the ratio of English words to Sinhala words in Romanized Sinhala text, providing valuable insights into the language blend and usage patterns within the author's writing. This feature enables the identification of the extent of code-switching and bilingual usage in the text. This was done using Swa-Bahsha an existing Transliteration system that maps a Romanized Sinhala piece of text to Sinhala, this acted as our mapping system to calculate the above features [8].

Additionally, the Levenshtein distance algorithm was employed to measure the text edit distance between similar Romanized Sinhala words across different chunks of text. This metric assists in evaluating the similarity and consistency of language usage, offering a quantitative assessment of linguistic variations and text coherence.

For example,
**Chunk 1**: *" warthamana janapathithuma wides sancharayak sadaha ada dina indiawa bala pitath uni "*

**Chunk 2**: *" wrthmna jnapthithuma widhes sncharyk sadha ada dina indiwa bala pitath wuni "*

The minimum number of single character edits is analysed for the computation of Levenshtein distance by considering insertion, deletion and substitution compared to the general representation of the context. In the above example, the Levenstein distance counts to be 12, showing that author is having specific writing pattern in his/her Romanized Sinhala texting. All the ad hoc representations of Romanized Sinhala writing are handled over here as a primary parameter to the prediction process.

*C. Similarity Computation*

The similarity computation involved comparing the feature vectors extracted from the suspect author's text and the anonymous text. For the English model, Euclidean distance metrics were applied to quantify the similarity between feature vectors. The Romanized Sinhala model utilized custom similarity measures tailored to the specific linguistic features extracted. Mainly , the English to Sinhala ratio of each chunk, text edit distance between similar romanized sinhala words, linguistic patterns in the writings are considered. This allowed for only the numerical representations to be used in training and allowed for the system to remain, non-domain and author-specific.

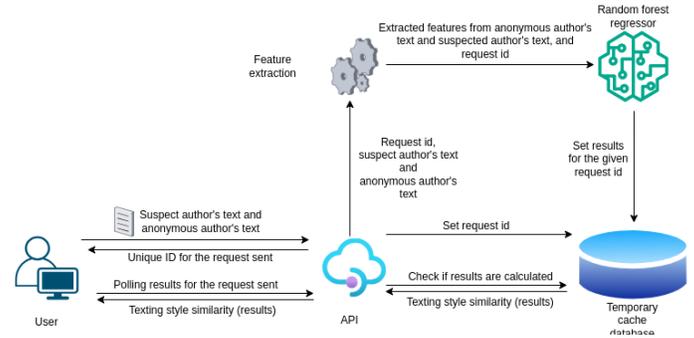

Fig. 1 The prototype diagram for the proposed system

*D. Random Forest Regressor for prediction*

The **Random Forest Regressor** was utilized in this study to predict the authorship similarity between pairs of texts, both in English and Romanized Sinhala, by analyzing feature-based similarity scores. Instead of directly analyzing each author's full set of texts, the model was trained to recognize patterns of similarity between two random pieces of text using extracted features such as word frequency, sentence structure, the ratio of Sinhala to English words (for Romanized Sinhala), and other stylometric elements. For each pair of texts, the similarity of these features was calculated, and the model learned to predict whether the two texts were likely authored by the same individual or different authors based on the patterns of similarity. The Random Forest Regressor was selected due to its ability to handle complex, feature-rich datasets and to generalize well across different text pairs. By building multiple decision trees, each looking at a subset of the feature similarities, the model was able to capture nuanced relationships between the texts [19].

The model's strength lies in combining the predictions of these trees to provide an accurate assessment of whether two texts share stylistic similarities that would indicate the same author. Additionally, Random Forest's robustness against overfitting was crucial in this context, where noisy and informal text data, such as text messages, blog posts and tweets, can lead to inconsistencies in patterns. The interpretability of the model allowed for insights into which features—such as the consistency of transliteration or the balance of English and Sinhala usage—were most important in determining whether two texts were authored by the same individual. Since the model focused on pairwise similarity rather than classifying individual authors' complete text sets, it was particularly effective in handling small to medium-sized datasets, especially when texts from the same or different authors were compared directly. This proved to be particularly helpful in the case of Romanized Sinhala where there was incredible difficulty finding data to be used for the corpus as Romanized Sinhala is only used in colloquial settings, with much of it's use being only on social media, from which getting data proves much ethical difficulty.

The Random Forest Regressor was employed to analyze the similarity scores between features in pairs of texts, making it an ideal choice for this specific project.

## IV. RESULTS AND DISCUSSION

The proposed methodology has been tested and evaluated using the custom-made dataset by the authors of this work. As this is a very novel area of research, the standard dataset for the evaluation of the authorship attribution in the literature for Romanized Sinhala and English are unavailable. Therefore, to perform the testing, we have used 80:20 split from the initial dataset to perform the evaluation for this phase.

*1) English*

```
              precision    recall  f1-score   support

           0       0.71      0.72      0.71        88
           1       0.69      0.68      0.68        81

    accuracy                           0.70       169
   macro avg       0.70      0.70      0.70       169
weighted avg       0.70      0.70      0.70       169
```

Fig. 2 The classification report for the English portion of the model
*Label 1 indicates that the 2 texts have the same author and Label 0 indicates that the 2 texts have different authors*

The model has a reasonable accuracy of 70%, this stands in similar standing to most another system in the domain for English Authorship Attribution despite using a relatively novel approach and having non-domain specificity. The model's balanced precision, recall, and F1 scores across classes suggest it effectively predicts both categories without much bias, which is crucial for tasks like authorship attribution. Its success, even without tuning to a specific domain, highlights strong generalization ability, making it versatile in various contexts. However, there is room for improvement, as 70% accuracy still leaves a margin for enhanced performance, particularly in high-stakes applications.

*2) Romanized Sinhala*

```
              precision    recall  f1-score   support

           0       0.86      0.96      0.91        56
           1       0.89      0.64      0.74        25

    accuracy                           0.86        81
   macro avg       0.87      0.80      0.83        81
weighted avg       0.87      0.86      0.86        81
```

Fig. 3 The classification report for the Romanized Sinhala portion of the model.

The report from Romanized Sinhala models suggests that the model accurately identifies the instances where an anonymous author's text and a suspected author's text are different from each other. However, there's a slight tendency that the model may also categorize texts by the same author as text from different authors. This could be as a result of a quite diverse way of representing each word in multiple ways in Romanized Sinhala.

The support for the model is also critical given that it has access to 56 works from different authors and only 25 works from the same authors. This imbalance in the dataset can contribute to the model's difficulty in accurately identifying texts from the same author. With a larger proportion of texts from diverse authors, the model may be able to better distinguish among different styles, resulting in the misclassification of texts from the same author as belonging to distinct authors. The limited sample size for the same author may not provide enough training data for the model to learn the unique features that consistently characterize an individual author's work.

As a summary of findings, the research questions are answered as follows.

(1) Can authorship attribution models be effectively developed for Romanized Sinhala text?

Authorship attribution models can be effectively developed for Romanized Sinhala text, as demonstrated. However, challenges are still present, particularly concerning the misclassification of texts from the same author, which may be influenced by the diverse representations of words in Romanized Sinhala.

(2) Can such a model generalize across different domains and linguistic structures?

Although the model has been demonstrated to function effectively across multiple domains, it has not yet been specifically tested on case texts from different domains. This limitation suggests that enhancements could be realized through cross-domain analysis, thereby increasing the model's robustness and adaptability to diverse linguistic structures and contexts.

Though our model presents a competitive result, there exist some limitations in the approach. The model is designed to work specifically with Romanized Sinhala and English texts where it cannot distinguish between the two if they are input into the wrong category. Its accuracy is highly dependent on the quality of the data provided by the user, meaning that poor or incorrect inputs can significantly affect its performance. Additionally, the model is ineffective when dealing with purposely obfuscated text, as it is not equipped to handle such distortions. Therefore, ensuring correct input and avoiding text manipulation is crucial for optimal results in this current work.

## V. CONCLUSION AND FUTURE DIRECTIONS

A robust non-domain specific author attribution system for English and Romanized Sinhala texts was developed. The system was to take inputs of the suspected author and the anonymous text and present a confidence score on if the specific text belongs to the suspected author. This system filled the gap in developing a pretrained model for assessing authorship of an unknown texts, not limited to a particular author or a domain.

This system also further addresses the need for a Romanized Sinhala model in Sri Lanka. This model was the 1st of its kind to be developed for any Romanized language, paving the way for more developments for authorship attribution models for other Romanized versions of languages like Arabic and Tamil.

Overall, this achieved what it set out to do, developing a robust authorship attribution model that is non-domain specific and can be applied to a plethora of authors in both English and Romanized Sinhala.

As a future direction to the study, cross domain analysis where the text from same author / suspect author is to come from 2 different domains could help with the model's robustness, thereby ensuring accurate authorship attribution across diverse contexts. Further research could focus on the investigation of transfer learning techniques, which may allow the model to leverage knowledge from related languages, thus enhancing its performance in languages/styles with fewer available resources like Romanized Sinhala [20]. Explainable AI methods could be implemented to clarify how the model makes its decisions. By highlighting the factors that affect authorship attribution, users can better understand the influence of specific features on predictions. This transparency will help build trust in the model and support future improvements [21]. The Public GitHub repository of the project- https://github.com/cipherdragon/SimpleAA